# Performance of Automatic De-identification Across Different Note Types

Nicholas Dobbins[1], David Wayne[2], Kahyun Lee[4], Özlem Uzuner, Ph.D.[4],
Meliha Yetisgen, Ph.D.[1,3]
[1]Biomedical and Health Informatics, [2]School of Medicine, [3]Linguistics,
University of Washington, Seattle, WA
[4] Information Sciences and Technology, George Mason University, Fairfax, VA

**Introduction**

Free-text clinical notes detail all aspects of patient care and have great potential to facilitate quality improvement and assurance initiatives as well as advance clinical research. However, concerns about patient privacy and confidentiality limit the use of clinical notes for research. As a result, the information documented in these notes remains unavailable for most researchers. De-identification (de-id), i.e., locating and removing personally identifying protected health information (PHI), is one way of improving access to clinical narratives. However, there are limited off-the-shelf de-identification systems able to consistently detect PHI across different data sources and medical specialties. In this abstract, we present the performance of a state-of-the art de-id system called NeuroNER[1] on a diverse set of notes from University of Washington (UW) when the models are trained on data from an external institution (Partners Healthcare) vs. from the same institution (UW). We present results at the level of PHI and note types.

**Dataset**

<u>UW-Dataset</u>: We created a dataset of 600K notes from patients who received treatment at University of Washington Medical Center and Harborview Medical Center between 2007 and 2017. From this dataset, we randomly sampled 1000 notes from 10 note types (100 notes each). The selected subset of 1000 notes contains a total of 1,890,849 tokens from the following note types: (1) admit notes, (2) discharge notes, (3) emergency department (ED) notes, (4) nursing notes, (5) pain management (PM) notes, (6) progress notes, (7) psychiatry notes, (8) radiology notes, (9) social work notes, and (10) surgery notes. The length of notes varies across note types (token count - avg: 1890.9; max: 2492.58 (progress notes); min: 948.55 (radiology); std-dev: 706.13). We annotated this set with 25 personal health identifiers (PHI). 8 graduate students from UW Biomedical Health Informatics departments and 2 medical students from UW School of Medicine completed the annotation. All notes were double-annotated and all conflicts were resolved. We grouped the 25 PHI under 6 PHI types (e.g., NAME: patient, doctor, user name). Table 1 includes the entity counts per PHI found in these notes.

| PHI Type | PHI | Admit Notes | Discharge | ED | Nursing | Pain Mgmt. | Progress | Psychiatry | Rad. | Social work | Surgery |
|---|---|---|---|---|---|---|---|---|---|---|---|
| NAME | Patient | 360 | 353 | 252 | 11 | 419 | 598 | 1351 | 24 | 844 | 31 |
|  | Doctor | 531 | 1505 | 2791 | 54 | 3475 | 1075 | 894 | 319 | 1334 | 177 |
|  | UserName | 0 | 0 | 1 | 0 | 3 | 0 | 0 | 0 | 0 | 0 |
| LOCATION | Room | 3 | 108 | 4 | 62 | 4 | 38 | 114 | 0 | 173 | 1 |
|  | Department | 81 | 662 | 72 | 5 | 222 | 282 | 89 | 170 | 196 | 38 |
|  | Hospital | 383 | 1189 | 327 | 192 | 995 | 694 | 609 | 128 | 1326 | 77 |
|  | Organization | 56 | 140 | 45 | 5 | 166 | 136 | 328 | 3 | 1493 | 0 |
|  | Street | 7 | 170 | 4 | 0 | 1110 | 117 | 10 | 84 | 36 | 0 |
|  | City | 116 | 162 | 45 | 0 | 939 | 136 | 93 | 21 | 313 | 5 |
|  | State | 41 | 94 | 6 | 1 | 833 | 85 | 28 | 21 | 183 | 1 |
|  | Country | 5 | 1 | 0 | 0 | 3 | 12 | 61 | 0 | 8 | 0 |
|  | Zip | 0 | 26 | 1 | 0 | 410 | 44 | 4 | 21 | 8 | 0 |
|  | Other | 19 | 23 | 6 | 0 | 588 | 40 | 13 | 0 | 153 | 0 |
| AGE | Age | 335 | 241 | 534 | 9 | 256 | 263 | 269 | 73 | 190 | 207 |
| DATE | Date | 4107 | 3463 | 2267 | 877 | 4566 | 4195 | 3052 | 683 | 903 | 2829 |
| CONTACT | Phone | 380 | 1786 | 259 | 6 | 384 | 287 | 369 | 120 | 1740 | 55 |
|  | Fax | 0 | 17 | 0 | 0 | 6 | 62 | 5 | 0 | 5 | 0 |
|  | Email | 0 | 2 | 0 | 0 | 0 | 7 | 2 | 0 | 46 | 0 |
|  | URL | 1 | 151 | 66 | 8 | 61 | 38 | 16 | 0 | 5 | 4 |
| IDs | MRN | 2 | 2 | 4 | 0 | 0 | 13 | 1 | 5 | 48 | 3 |
|  | ID-Number | 14 | 105 | 288 | 0 | 76 | 10 | 6 | 5 | 26 | 47 |
|  | Account | 2 | 0 | 0 | 0 | 0 | 0 | 0 | 0 | 0 | 0 |
|  | Health Plan | 0 | 8 | 0 | 0 | 0 | 5 | 9 | 0 | 119 | 0 |
| PROFESSION | Profession | 84 | 61 | 13 | 0 | 129 | 63 | 85 | 0 | 80 | 0 |
|  | Total: | 6527 | 10269 | 6985 | 1230 | 14645 | 8200 | 7408 | 1677 | 9229 | 3475 |

**Table 1.** Annotation statistics for PHI entities across different not types.

As can be seen from the table, each note type has different distributions of PHI content. Pain management, discharge, and social work notes are the top three note types for PHI content density. Radiology and nursing notes include significantly fewer PHI when compared to other note types. There is also variance in the density of PHI types across note types. For example, patient name is mentioned more frequently in psychiatry and to a lesser extent social work notes, while age, medical record number, and profession are rarely mentioned in nursing and radiology notes.

<u>i2b2 Dataset</u>: We used the 976 longitudinal notes, including doctor's notes, discharge summaries, doctor-patient correspondence, and lab results, available in the training set of 2014 i2b2/UTHealth shared task[2] as external data to measure the domain adaptability between Partners Healthcare and UW. Our UW data were annotated according to the annotation guidelines associated with the 2014 i2b2/UTHealth shared task data, providing two consistently annotated data sets for de-identification across institutions.

**Methods**

We define de-identification as a named entity recognition (NER) task. We evaluated the performance of an NER system, called NeuroNER[1], on our dataset. NeuroNER uses long short-term memory (LSTM)-based recurrent neural networks (RNN) for non-overlapping label prediction and achieves state-of-the-art performance on a number of tasks.

In our experimental setup, we split the 1000 annotated UW notes with a 4:1 ratio (training set: 800, test set: 200). We ran the following experiments: (1) trained and tuned on external data (i2b2 de-id dataset[2]) and tested on UW-test set, (2) trained and tuned on in-house data (UW-training set) and tested on UW-test set, and (3) trained and tuned on external and in-house data (i2b2 de-id dataset + UW-training set) and tested on UW-test set. We compared performance differences across models at the PHI and note type levels.

**Results and Conclusion**

Table 2 presents extraction results at the PHI type level. Overall training only with i2b2 notes provides a reasonable starting point for de-identification on UW data. However, performance is significantly better when the models were trained with the UW training set. The combined i2b2 + UW-training set achieved the best performance across 5 of 6 PHI types, with a mean per-label F1-score improvement of 1.2 over training with UW notes alone. Among PHI types, dates showed the best performance, with 97.5 F1-score, while professions showed poorest performance with 50.0 F1-score. We believe the comparatively high performance of date identification can be attributed to (1) the large number of date instances w.r.t. other labels, and (2) the relatively small number of date patterns (e.g., '2019-4-8', or 'March 8$^{th}$'). For the worst-performing type, profession, we believe the main reason for its poorer performance is due to a smaller number of training samples.

Table 3 presents F1-scores at the note type level. Among note types, the addition of UW training notes to i2b2 notes boosted F1-scores by a mean 12.3 (std-dev: 5.8). Radiology notes showed the greatest improvement between training with i2b2 notes versus UW notes (+27.5) – i2b2 data contained no radiology notes. Across nearly all note types, the relatively large number of dates raises the overall F1 scores significantly, despite comparatively lower performance of other types, such as professions. Our results suggest that training NeuroNER using multi-institutional corpora for de-identification tasks can improve identification of certain types of PHI.

| Training set | Name | Location | Age | Date | Contact | IDs | Profession |
|---|---|---|---|---|---|---|---|
| I2b2 | 75.9 | 59.8 | 81.0 | 92.2 | 66.5 | 40.2 | 25.0 |
| UW | 91.0 | **83.1** | 89.7 | 97.4 | **86.9** | 82.8 | 45.6 |
| I2b2 + UW | **92.6** | 83.0 | **91.4** | **97.5** | **87.0** | **83.6** | **50.0** |

**Table 2.** De-identification performance (F1-score) across different types of PHI.

| Training set | Admit | Discharge | ED | Nursing | Pain Mgmt. | Progress | Psychiatry | Rad. | Social work | Surgery |
|---|---|---|---|---|---|---|---|---|---|---|
| I2b2 | 88.6 | 82.41 | 83.1 | 90.0 | 82.5 | 82.9 | 82.6 | 70.2 | 79.6 | 87.2 |
| UW | 97.2 | 94.4 | 93.7 | 97.7 | 95.3 | 93.6 | 95.9 | 97.7 | 90.8 | 95.8 |
| I2b2 + UW | 96.7 | 92.8 | 95.6 | 95.9 | 97.4 | 94.2 | 95.5 | 97.5 | 91.0 | 95.6 |

**Table 3.** De-identification performance (F1-score) across different note types.

**Acknowledgements**

This study was supported by the National Library of Medicine under Award Number R15LM013209 and by the National Center For Advancing Translational Sciences of National Institutes of Health under Award Number UL1 TR002319. Experiments were run on computational resources from the UW Department of Radiology.